# BotanicGarden: A High-Quality Dataset for Robot Navigation in Unstructured Natural Environments

Yuanzhi Liu[1], Yujia Fu[1,†], Minghui Qin[1,†], Yufeng Xu[1,†], Baoxin Xu[1], Fengdong Chen[2], Bart Goossens[3,*], Poly Z.H. Sun[4], Hongwei Yu[5], Chun Liu[6], Long Chen[7], Wei Tao[1], Hui Zhao[1,*]

***Abstract*—The rapid developments of mobile robotics and autonomous navigation over the years are largely empowered by public datasets for testing and upgrading, such as sensor odometry and SLAM tasks. Impressive demos and benchmark scores have arisen, which may suggest the maturity of existing navigation techniques. However, these results are primarily based on moderate structured scenario testing. When transitioning to challenging unstructured environments, especially in GNSS-denied, texture-monotonous, and dense-vegetated natural fields, their performance can hardly sustain at a high level and requires further validation and improvement. To bridge this gap, we build a novel robot navigation dataset in a luxuriant botanic garden of more than 48000m$^2$. Comprehensive sensors are used, including Gray and RGB stereo cameras, spinning and MEMS 3D LiDARs, and low-cost and industrial-grade IMUs, all of which are well calibrated and hardware-synchronized. An all-terrain wheeled robot is employed for data collection, traversing through thick woods, riversides, narrow trails, bridges, and grasslands, which are scarce in previous resources. This yields 33 short and long sequences, forming 17.1km trajectories in total. Excitedly, both highly-accurate ego-motions and 3D map ground truth are provided, along with fine-annotated vision semantics. We firmly believe that our dataset can advance robot navigation and sensor fusion research to a higher level.**

Website: https://github.com/robot-pesg/BotanicGarden

## I. Introduction

MOBILE robots play a crucial role in today's social development and productivity evolution. Over the years, with the rapid progress of autonomous navigation, various applications have emerged, such as robotaxi, unmanned logistics, service robots, and more [1]. Meanwhile, existing algorithms begin to saturate the benchmarks, which may suggest that current navigation techniques have achieved a maturity in moderate and structured scenarios. However, robots often need to perform more complex tasks and work in unstructured environments, which consequently imposes higher demands on the capabilities and robustness of navigation systems.

Modern navigation techniques such as Sensor Odometry (SO) and Simultaneous Localization and Mapping (SLAM) [2] are indeed highly dependent on good scene compatibility and positioning aids to avoid tracking losses and cumulative drift. In well textured and structured environments, both vision- and LiDAR-based navigation methods can operate reliably by integrating inertial sensors and external positioning signals.

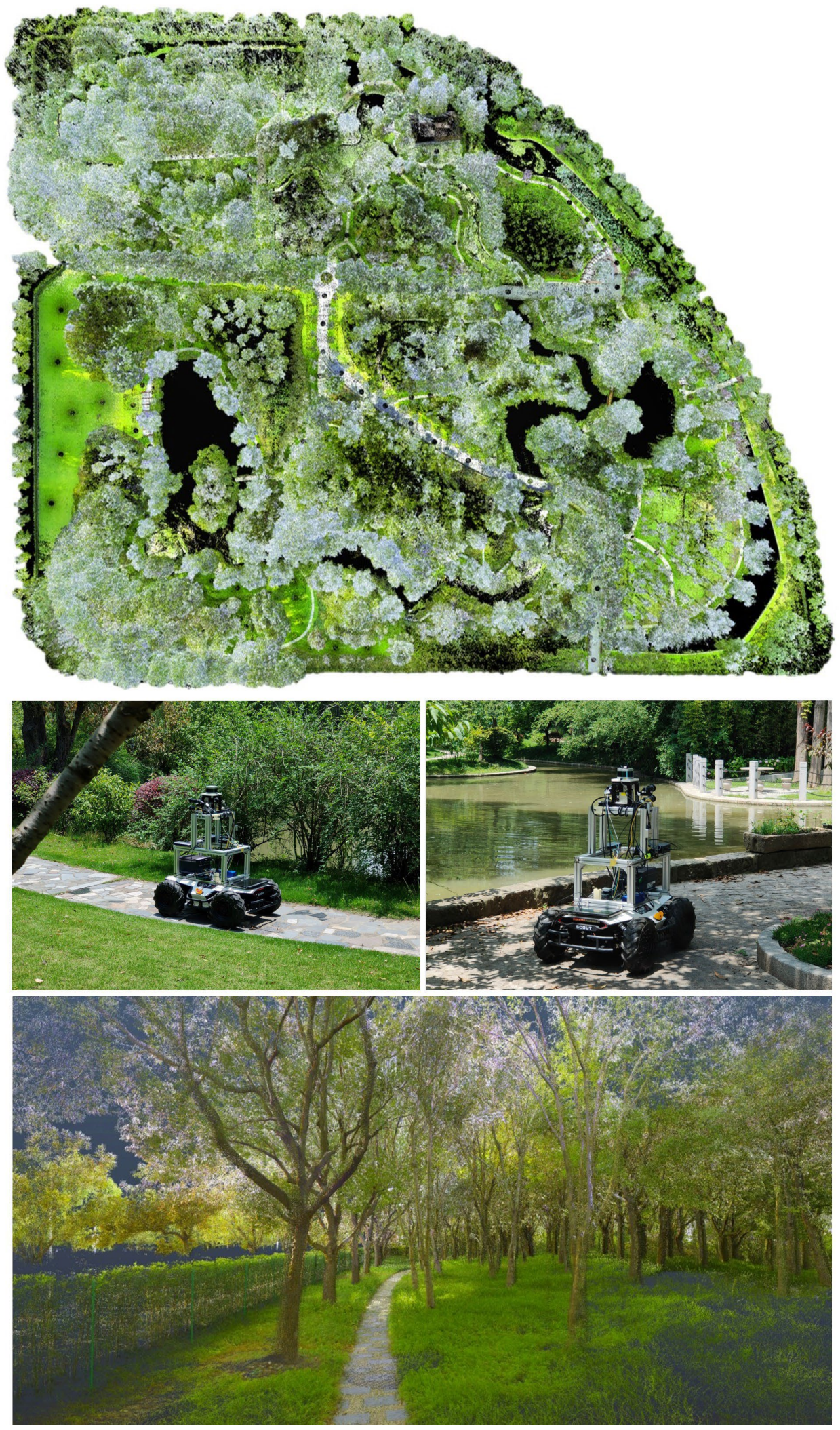

Fig.1. **Top**: A bird view of the 3D survey map of *BotanicGarden*; **Middle**: The robot is walking through the narrow path and riverside; **Bottom**: A detailed view of the 3D map in GNSS-denied thick woods.

Manuscript received: October 15, 2023; Revised December 31, 2023; Accepted January 17, 2024. This paper was recommended for publication by Editor Javier Civera upon evaluation of the Associate Editor and Reviewers' comments. This work was supported by the National Key R&D Program of China under Grant 2018YFB1305005. *(†Equal second contribution: Yujia Fu; Minghui Qin; Yufeng Xu) (*Corresponding author: Bart Goossens; Hui Zhao.)*

[1] Yuanzhi Liu, Yujia Fu, Minghui Qin, Yufeng Xu, Baoxin Xu, Wei Tao, and Hui Zhao are with School of Sensing Science and Engineering, Shanghai Jiao Tong University, Shanghai 200240, China (e-mail: lyzrose@sjtu.edu.cn; yujiafu@sjtu.edu.cn; sheeping-dragon@sjtu.edu.cn; xuyufeng@sjtu.edu.cn; 2583035553@sjtu.edu.cn; taowei@sjtu.edu.cn; huizhao@sjtu.edu.cn).

[2] Fengdong Chen is with School of Instrumentation, Harbin Institute of Technology, Harbin 150001, China (e-mail: chenfd@hit.edu.cn).

[3] Bart Goossens is with imec-IPI-Ghent University, 9000 Gent, Belgium (e-mail: bart.goossens@ugent.be).

[4] Poly Z. H. Sun is with School of Mechanical Engineering, Shanghai Jiao Tong University, Shanghai 200240, China (e-mail: zh.sun@sjtu.edu.cn).

[5] Hongwei Yu is with Chinese Aeronautical Radio Electronics Research Institute, Shanghai 200233, China (e-mail: hwyu615@163.com).

[6] Chun Liu is with College of Surveying and Geo-Informatics, Tongji University, Shanghai 200092, China (e-mail: liuchun@tongji.edu.cn).

[7] Long Chen is with the Institute of Automation, Chinese Academy of Sciences, Beijing 100190, China (e-mail: long.chen@ia.ac.cn).

However, in problematic unstructured scenarios involving GNSS denial, textural monotonicity, and especially within dense-vegetated natural fields, their performances can hardly sustain at a high level and necessitate further validation.

As is well known, due to the costly hardware and complicated experiments, robot navigation research relies heavily on publicly available datasets for testing and upgrading [3]. The most famous resources, including KITTI [4], TUM-RGBD [5], and EuRoC [6], have become the indispensable references in today's algorithm developments. Other newer datasets such as NCLT [7], Oxford RobotCar [8], Complex Urban [9], Newer College [10], and 4-Seasons [11] also complement a wide scene variety. However, such datasets are mainly with urbanized and indoor environments, which cannot serve as qualified benchmarks for the aforementioned problematic scene settings. This motivates us to build a novel dataset in unstructured natural environments to further promote research in robot navigation.

In this paper, we introduce a high-quality robot navigation dataset which is collected in a luxuriant botanic garden of over 48000m$^2$. An all-terrain robot, equipped with strictly integrated stereo cameras, LiDARs, IMUs, and wheel odometry, traverses diverse natural areas including dense woods, riversides, narrow trails, bridges, and grasslands, as depicted in Fig. 1. Here GNSS cannot work reliably due to the block of thick vegetations, and the repetitive green features and unstructured surroundings may also shake the performance of motion and recognition modules. The work most similar to ours could be Montmorency [12], while it focuses more on LiDAR mapping, lacking in sensors variety, scene scale and diversity, and authentic ground truth. Our main contributions are as follows:

- We build a novel multi-sensory dataset in an over 48000m$^2$ botanic garden with 33 long and short sequences and 17.1km trajectories in total, containing dense and diverse natural elements that are scarce in previous resources.
- We employed comprehensive sensors, including high-res and high-rate stereo gray and RGB cameras, spinning and MEMS 3D LiDARs, and low-cost and industrial-grade IMUs, supporting a wide range of applications. By elaborate development of the system, we have achieved highly-precise hardware-synchronization. Both the availability of sensors and sync-quality are at the top-level in this field.
- We provide both highly-precise 3D map and trajectories ground truth by dedicated surveying works and advanced map-based localization algorithm. We also provide dense vision semantics labeled by experienced annotators. This is the first field robot navigation dataset that provides such all-sided and high-quality reference data.

## II. Related Works

### A. SO/SLAM-based Navigation

Traditional navigation systems are typically achieved with GNSS (Global Navigation Satellite System), and filtered with inertial data. GNSS can provide drift-free global positioning at meters level, while inertial data are in duty of attitude and can boost the frequency to more than 100Hz. However, as is well known, GNSS requires an open-sky to locate reliably, while is unworkable indoors and is out of precision in denied outdoor areas such as urban canyon, tunnels, and forests. These failure cases motivate the developments of modern SO/SLAM-based navigation which employ vision and LiDAR as centric sensors. SO is the process of tracking an agent's location incrementally over time, with perception and navigation sensors. It has been widely researched over the years, forming mature implementations such as Visual and Visual-Inertial Odometry (VO/VIO), which are compact and computationally lightweight. As an extension of SO, SLAM is a process of building a map of the environments while simultaneously keeping the track of the agent's locations within it. Compared with SO, SLAM could be more accurate and robust: by loop closure corrections, SLAM is able to optimize the map and path to bound cumulative drifts; and it is also possible to re-localize after tracking losses by searching the base map. Famous SO/SLAM frameworks include VINS-Mono [13], ORB-SLAM [14], [15], LOAM and its extensions [16], [17], *etc*. According to the benchmark results, state-of-the-art methods exhibit good performance in structured environments and can handle occasional challenges. However, their robustness in complex unstructured scenarios characterized by dense natural elements and monotonous textures remains questionable and necessitates further validation.

### B. Representative Datasets

Over the past two decades, the field of mobile robotics has witnessed the introduction of numerous publicly available datasets, mainly consisting of structured environments such as urban, campus, and indoor scenarios. Among the earliest efforts, the most notable presented datasets include MIT-DARPA [18], Rawseeds [19], and KITTI [4]. These datasets offered a comprehensive range of sensor types and accurate ego-motion ground truth derived from D-GNSS systems. During this early phase, the main objective of these datasets was to fulfill basic testing and validation requirements. As a result, the collection environments were intentionally designed to be relatively simple. However, exactly due to the idealistic illuminations, weathers, and static scene layouts, these datasets have received concerns for being too ideal for algorithm assessments [4].

To complement previous datasets with a greater emphasis on real-life factors, significant efforts have been made in the subsequent years. On the one hand, several long-term datasets have been proposed, including NCLT [7], Oxford RobotCar [8], KAIST Day/Night [20], and 4-Seasons [11], incorporating diverse temporal variations, weather conditions, and seasonal effects. On the other hand, to address the need for more complex and dynamic environments, ComplexUrban [9] and UrbanLoco [21] were developed. ComplexUrban focused on metropolitan areas in South Korea, while UrbanLoco covered cities in Hong Kong and San Francisco, bringing in challenging features like urban canyon, dense buildings, and congested traffics. Throughout this stage, datasets have played a crucial role in pushing the boundaries of algorithms, aiming to enhance their robustness for real-world applications.

Many indoor and 6-DoF datasets also exist. Famous repositories include TUM-RGBD [5], EuRoC [6], TUM-VI [22], and more, which significantly promote the research of visual and visual-inertial navigation systems (VINS). Besides, in recent years, high-quality multi-modal datasets were also continuously emerging, such as OpenLORIS [23], M2DGR [24], Newer College [10], and Hilti SLAM [25]. These datasets encompass a wide range of real-life challenges, providing valuable opportunities for algorithm validation and improvement.

Up to the present, there is a relatively abundant availability of datasets in structured environments, which have become increasingly comprehensive and challenging. However, while existing algorithms have shown promising performance in such scenarios, due to the wide variation in scene patterns, their capabilities in unstructured environments remains questionable and necessitate concrete and targeted validation.

TABLE I. COMPARISON OF DIFFERENT NAVIGATION DATASETS

| Dataset | Environment | | Platform | Camera | Stereo Vision | | | | 3D LiDAR | IMU | Sync | GT-Pose[1] | GT-Map[2] | Semantic |
|---|---|---|---|---|---|---|---|---|---|---|---|---|---|---|
| | Scene | Type | | | Gray | RGB | Resolution | Rate | | | | | | |
| KITTI [4] | Urban | Struct | Vehicle | ✓ | ✓ | ✓ | 1392×512 | 10 | ✓ | ✓ | Hw/Sw | D-GNSS/INS(✓) | - | Dense |
| TUM-RGBD [5] | Indoor | Struct | Handheld Robot | ✓ | - | - | - | - | - | ✓ | Sw | MoCap[3](✓) | - | - |
| EuRoC [6] | Indoor | Struct | Drone | ✓ | ✓ | - | 752×480 | 20 | - | ✓ | Hw | LasTrack[4](✓) MoCap(✓) | Scanner(✓) | - |
| DARPA [18] | Urban | Struct | Vehicle | ✓ | - | - | - | - | ✓ | ✓ | Sw | D-GNSS/INS(✓) | - | - |
| NCLT [7] | Campus Building | Struct | Robot | ✓ | - | - | - | - | ✓ | ✓ | Hw/Sw | D-GNSS/INS(✓) SLAM | - | - |
| RobotCar [8] | Urban | Struct | Vehicle | ✓ | - | ✓ | 1280×960 | 16 | ✓ | ✓ | Hw/Sw | D-GNSS/INS(✓) | - | - |
| M2DGR [24] | Campus Building Lab | Struct | Robot | ✓ | - | ✓ | 1280×1024 | 15 | ✓ | ✓ | Sw | D-GNSS/INS(✓) LasTrack(✓) MoCap(✓) | - | - |
| Rellis-3D [36] | Off-road | Unstruct | Robot | ✓ | - | ✓ | 800×592 | 10 | ✓ | ✓ | Hw/Sw | GNSS/INS | - | Dense |
| TartanDrive [37] | Off-road | Unstruct | Vehicle | ✓ | - | ✓ | 1024×512 | 20 | ✓ | ✓ | Hw/Sw | D-GNSS/INS(✓) | - | - |
| FinnForest [38] | Forest | Unstruct | Vehicle | ✓ | - | ✓ | 1920×1200 | 40 | - | ✓ | Hw | D-GNSS/INS(✓) | - | - |
| Wild-Places [40] | Forest | Unstruct | Handheld | ✓ | - | - | - | - | ✓ | ✓ | Hw/Sw | SLAM | SLAM | - |
| Montmorency [12] | Forest | Unstruct | Robot | ✓ | - | - | - | - | ✓ | ✓ | Sw | SLAM | SLAM | - |
| **Ours** | **Natural** | **Unstruct** | **Robot** | ✓ | ✓ | ✓ | **1920×1200** | **40** | **vlp16+livox** | ✓ | **Hw** | **GT-map ICP(✓)** | **Scanner(✓)** | **Dense** |

[1] GT-pose: ground truth pose. [2] GT-map: ground truth map. [3]MoCap: Motion Capture System. [4]LasTrack: Laser Tracker. ✓ denotes the ground truth data are authentic.

### C. Datasets in Unstructured Environments

Unstructured environments refer to scenarios that lack clear, regular, or well-defined features. In such scenarios, there is typically a lack of apparent motifs or geometric shapes, which raise great difficulty for robotics algorithms to recognize and track. Datasets in unstructured environments typically involve those collected in sandy and rocky fields, undergrounds, rural areas, rivers, and scenes rich in natural elements (forests, wilds, and diverse vegetations). Different from datasets in structured environments, efforts paid to unstructured scenarios are relatively less. An overview of existing works is given below.

For datasets with sandy and rocky scenarios, Furgale *et al.* [26] created a long-range robot navigation dataset with stereo cameras on Devon Island; Vayugundla *et al.* [27] recorded two sequences on Mount Etna with stereo vision, IMU, and odometry sensors; Hewitt *et al.* [28] collected a dataset in Katwijk beach with a wide array of high-quality sensors; and Meyer *et al.* [29] recorded diverse visual-inertial sequences in the Moroccan desert. These datasets were challenging for vision methods mainly due to the monotonous texture of the scenes.

For underground environments, Leung *et al.* [30] collected a 2km sequence in a large mine of Chile, and Rogers *et al.* [31] created a grand dataset within a huge tunnel circuit. In such cases, the challenge mainly lies on the absence of GNSS, where the robots may rely solely on ego-sensors for navigation.

For scenes of rural areas and rivers, Chebrolu *et al.* [32] and Pire *et al.* [33] collected various sequences in croplands; and VI-Canoe [34] and USVInland [35] respectively built a dataset in rural rivers and inland waterways. They introduced challenges related to the lack of distinct features and interference caused by water and surrounding vegetations.

For scenes rich of natural elements, which is also the scope of our work, Rellis-3D [36] and TartanDrive [37] focused on multi-modal datasets in off-road terrains, and FinnForest [38] recorded diverse visual-inertial sequences along wide roads in a large forest. These datasets intentionally incorporated challenges related to monotonous textures and lack of structural cues, yet they still managed to secure reliable GNSS signals, which may not represent the most demanding case in such scope. Towards inner and denser natural spaces, where GNSS cannot work reliably, RUGD [39], Montmorency [12], and Wild-Places [40] collected diverse data in thick vegetations. However, exactly due to the blockage of GNSS signal, they failed to provide authentic ground truth for ego-motion: Montmorency and Wild-Places employed SLAM algorithms to estimate the trajectories, while RUGD did not release any trajectory data in the original paper. As a result, they were better suited for the validation of scene perception and place recognition tasks, rather than for strict-sensed robot navigation, which primarily focuses on state estimation. This serves as the motivation of our paper.

### D. Discussions

In summary, there is a significant gap between unstructured and structured environments in terms of scene patterns, which poses much severer challenges for navigation algorithms. However, existing datasets still have notable limitations in this regard, particularly in environments with dense natural elements and degraded GNSS services, where the obtainment of ground truth remains a problem. This paper fills the gap by introducing a novel high-quality dataset in a luxuriant botanic garden. Table I compares our work with key state-of-the-arts and several highly-relevant unstructured counterparts, showing that our sensors availability, time-sync, and ground truth quality are all at the top-level in this field. We thus believe that our work will be extremely beneficial for mobile robotics community.

## III. THE BOTANIC GARDEN DATASET

### A. Acquisition Platform

To cope with the complex field environments, we employ an all-terrain wheeled robot Scout V1.0 from AgileX for data collection. It works in a powerful 4-wheel-drive (4WD) differential mechanism, which can ensure high driving robustness and obstacle crossing ability in fields and wilds. Each wheel contains a 1024-line encoder to provide ego-motions, and we have developed a set of corresponding programs to calculate the robot dead-reckoning odometers. To ensure a low latency communication, the robot is configured to link with the host via a high-speed CAN bus at 500kbps, which can lower the transmission time to less than 1ms. Besides, the host controller is performed by an Intel NUC11 running with a Real-Time Linux kernel[1] to minimize the clock jitter and data buffer time. We have customized the NUC to support dual-Ethernet with Precision Time Protocol (PTP[2], also known as IEEE1588) capability, which is able to be synchronized with other devices at sub-µs accuracy.

[1] https://wiki.linuxfoundation.org/realtime/start
[2] https://standards.ieee.org/ieee/1588/4355/

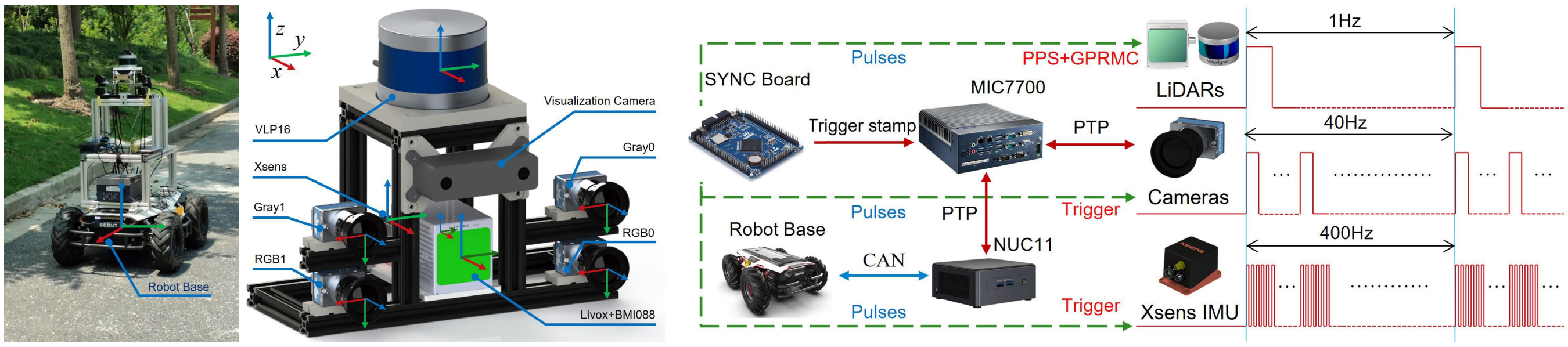

Fig.2. **Left**: The robot platform design and its base coordinate; **Middle**: The multi-sensory system and the corresponding coordinate (the camera below the VLP16 is only for visualization usage, thus is not annotated); **Right**: The synchronization system of the whole platform.

TABLE II. SPECIFICATIONS OF SENSORS AND DEVICES

| Sensor/Device | Model | Specification |
|---|---|---|
| Gray Stereo | DALSA M1930 | 1920*1200, 2/3", 71°×56°FoV, 40Hz |
| RGB Stereo | DALSA C1930 | 1920*1200, 2/3", 71°×56°FoV, 40Hz |
| Spinning LiDAR | Velodyne VLP16 | 16C, 360°×30°FoV, ±3cm@100m, 10Hz |
| MEMS LiDAR | Livox AVIA | 70°×77°FoV, ±2cm@200m, 10Hz |
| D-GNSS/INS | Xsens Mti-680G | 9-axis, 400Hz, GNSS not in use |
| Consumer IMU | BMI088 | 6-axis, 200Hz, Livox built-in |
| Wheel Odometry | Scout V1.0 | 4WD, 3-axis, 200Hz |
| 3D Survey Scanner | Leica RTC360 | 130m range, 1mm+10ppm accuracy |

On top of the robot chassis, we design a set of aluminum profiles to carry the batteries, computers, controllers, sensors, and the display, as illustrated in Fig. 2. The computer used for data collection is an Advantech MIC-7700 Industrial PC assembled with a PCIE expansion module. It houses an Intel Core i7-6700TE 4C8T processor running with Ubuntu 18.04.1 LTS and ROS Melodic systems. A total of 8 USB 3.0, 10 GigE, and a set of GPIO and serial ports are available. All the GigE ports supports PTP, available for precise time synchronization. For high-speed data logging, 2×16GB DDR4 memories (dual-channel) and a 2TB Samsung 980 Pro NVME SSD (of 3-bit MLC, over 1.5GB/s sequential writes throughout the whole storages) are equipped for real time database. To ensure full communication bandwidth, both the GigE cards (for sensor streaming) and the SSD are fastened to the PCIE slots that directly linked to the CPU. Benefiting from our elaborate development, this system can record over 500MB/s data stream without losing a single piece of image, which is a common issue in many other datasets.

## B. Sensor Setup

Our dataset focuses on robot navigation research based on conventional mainstream sensor modalities and their fusions. To this end, we have employed comprehensive sensors including stereo Gray and RGB cameras, spinning and MEMS 3D LiDARs, and low-cost and industrial-grade IMUs. Their specifications are as listed in Table II. All the sensors are accurately mounted on a compact self-designed aluminum carrier with precise 3D printing fittings, as shown in Fig. 2.

The stereo sensors are composed of two grayscale and two RGB cameras with a baseline of around 255mm. To facilitate research on robotic vision, we have chosen models from Teledyne DALSA with both high rate and resolution: M1930 and C1930, working at 1920×1200 and 40fps in our configuration. The CMOS used for the cameras is the PYTHON 2000 from ONSemi with 2/3" format and 4.8μm pixel size, which has a good performance under subnormal illuminations. However, this sensor in its nature has very strong infrared response, thus we have customized IR-cutoff filters of 400-650nm to exclude the side-effects on white-balance and exposure. The cameras use GPIO as external trigger, and GigE for data streaming, which also supports PTP synchronizations. The attended lens for imaging is Ricoh's CC0614A (6mm focus and F1.4 iris), which has been adjusted to 5-10m clear view to fit the scene.

To support different testing demands, 2 LiDARs are used in collection: Velodyne VLP16 and Livox AVIA. VLP16 is a cost-effective spinning LiDAR with a 360°×30° Field-of-View (FoV), suitable for ground robotic navigation tasks. AVIA is a MEMS 3D LiDAR with non-repetitive 70°×77° circular FoV, thus is more suitable for dense mapping and sensor-fusion with co-heading cameras. They are both configured to scan at 10Hz, and can be synchronized via pulse-per-second (PPS) interface.

For inertial sensors, we provide a low-cost BMI088 IMU (200Hz) and an industrial-grade Xsens Mti-680G D-GNSS/INS system (IMU@400Hz, GNSS not in use) for comparison usage. BMI088 is built-in and synchronized with AVIA LiDAR, and Xsens supports external trigger via pulse rising edges.

## C. Time Synchronization

In a precise robot system with rich sensors and multi-hosts, time synchronization is extremely vital to eliminate perception delay and ensure navigation accuracy. Towards a high-quality dataset, we have taken very special cares on this problem. Our synchronization is based on a self-designed hardware Trigger and Timing board and a PTP-based network, as illustrated in Fig. 2. The Trigger and Timing board is implemented by a compact STM32 MCU. It is programmed to produce three channels of pulses 1Hz-40Hz-400Hz in the same phases. The 1Hz channel (PPS) is used for the synchronization of VLP16 and AVIA accompanied with GPRMC signals: Every time the rising edge arrives, LiDAR immediately clears its internal sub-second counter, thus all the point clouds in the subsequent second can be timed cumulatively based on PPS arrival, which will then be appended with UTC integer time by GPRMC. The 40Hz signal is used to trigger the cameras, when a rising edge arrives, the global shutter will immediately start exposure until reaching a target gain, and the image timestamp is acquired by adding half the exposure time to the trigger stamp. The 400Hz signal is used for triggering the Xsens IMU: Xsens has its own internal clock, and when the rising edge arrives, Xsens will be triggered an external interruption thus feedbacks its exact time, then our program can bridge a transform thus stamp the neighboring sample instance. The UTC time is maintained by MCU based on its onboard oscillator. Note that, to maintain the timing smoothness, we will never interrupt the MCU clock during the collections, instead, an UTC stamp will be conferred at the begin of each course-day via NTP or GNSS timing. So far, the LiDAR-vision-IMU chain has been fully synchronized in hardware. With a sub-μs level triggering consistency between the sensors (see Fig. 3), a high sync-precision should be obtained.

The PTP-based network is designed for synchronizing multi-hosts and capturing trigger events, thus the wheel odometry can be aligned to an identical timeline with other sensors. Our network frame is built based on LinuxPTP[3] library. We assign MIC-7700 as grand master, and DALSA cameras and NUC11 are configured as slaves. When the synchronization starts, the slaves will keep exchanging sync-packets with the master, and to ensure the smoothness of local clocks, we have not directly compensated the offsets, while instead employ a PID mechanism to adjust the time and frequency. During the data collection, once the camera is triggered, it will report its timestamp of PTP clock, and based on the MCU trigger stamp, our software will bridge a relation thus transforms the wheel odometry from PTP to MCU timeline. Here although the PTP network and the real-time kernel are used, there exists a latency from the CAN bus of around 1ms, which has been compensated in advance.

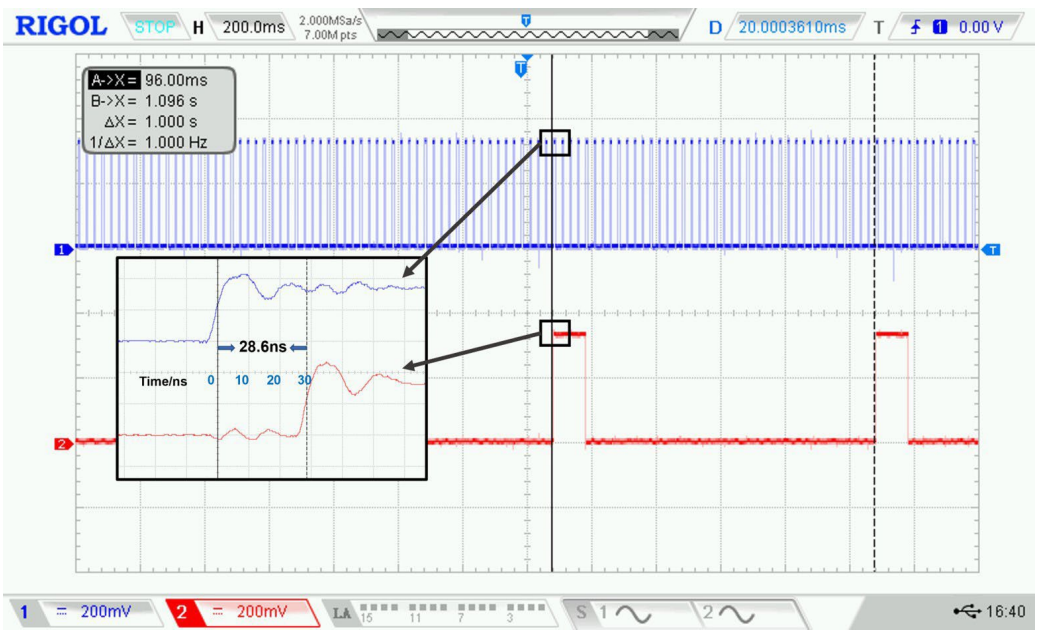

Fig.3. The hardware triggered 1Hz-40Hz pulses and rising edge offsets, indicating a high sync-precision at sensors side.

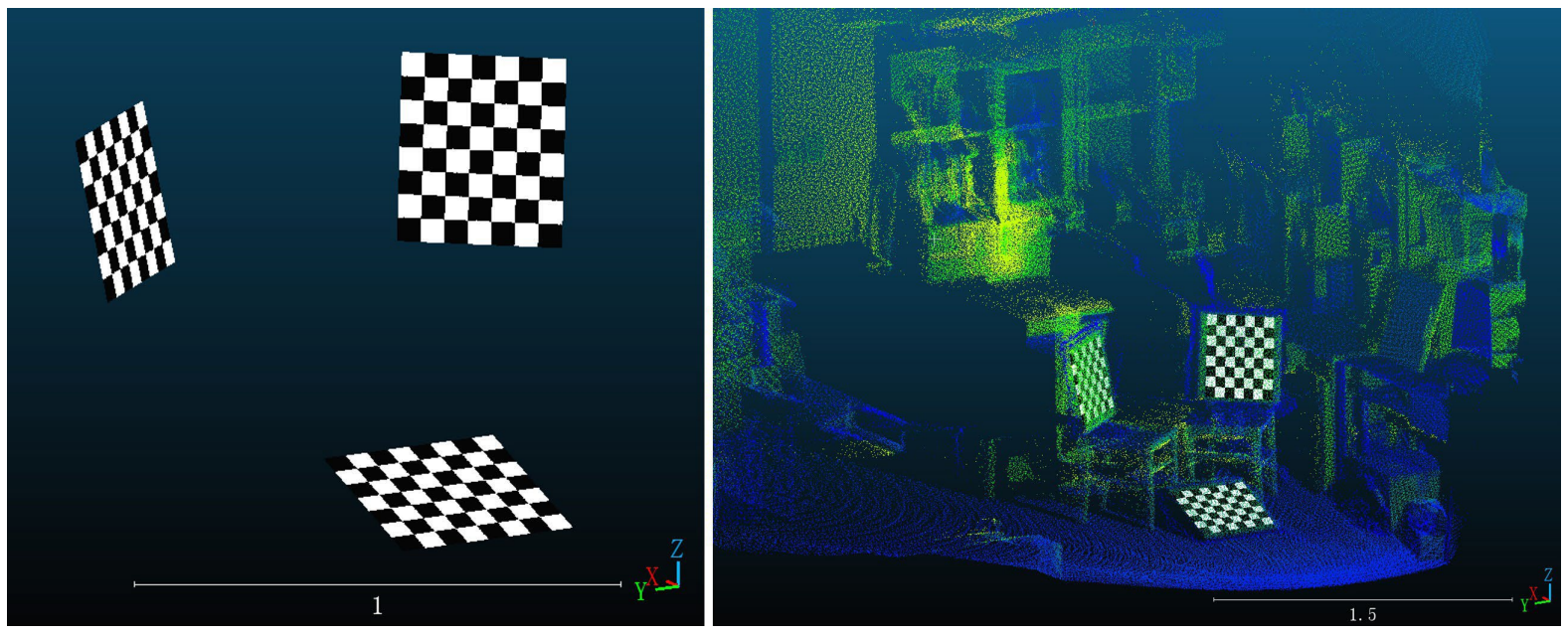

Fig.4. Camera-LiDAR calibration: **Left**: The checkerboards reconstructed by stereo vision; **Right**: The registration of vision 3D reconstruction model and LiDAR point cloud.

## *D. Spatial Calibration*

Spatial calibration, both for intrinsic and extrinsic parts, is a prerequisite for algorithm development. We ensure calibration quality through careful error evaluation and manual verification of the results. Note that, the calibration is performed based on the mounting positions of the sensors on the robot, as they have already been well assembled according to the CAD designs.

1) Camera calibration: For camera intrinsics and extrinsics calibration, we choose the Matlab camera calibration toolbox[4], which uses an interactive engine for inspecting the errors and filtering the qualified instances. Considering the standard lens FoV, we choose Pinhole imaging model ($f_x$, $f_y$, $c_x$, $c_y$) and a 4th degree polynomial Radial distortion model ($k_1$, $k_2$, $p_1$, $p_2$) for intrinsics. The calibration is conducted by manually posing a large checker board (11×8, 60mm/square) at different distances and orientations in front of the cameras. To avoid possible motion blur, the exposure has been controlled to ≤10ms, and we finally achieve less than 0.1pixels mean reprojection error in all the 4 cameras. Furthermore, based upon these intrinsics, the extrinsics are finely calculated via joint optimizations, and we have checked the epipolar coherence for a verification.

2) Camera-IMU calibration: The extrinsics between cameras and IMUs are determined using the famous Kalibr[5] toolbox. Thanks to our specially-designed detachable sensors suite, we are able to handheld it for 6-DoF movements. Before running the joint calibration, we have recorded 20 hours of IMU sequences to identify their intrinsics (noise densities and random walks of the accelerometers and gyroscopes). During the calibration, we use a 6×6 Aprilgrid as stationary target and properly move the sensor suite to excite all IMU axes. To avoid excessive motion blur, we have conducted the calibration in good lights and limited the exposure to ≤10ms. Note that, this joint calibration can also output time offset, whereas, as the sensors have already been hardware-synced, thus to avoid the side effects, this workflow is limited to camera-IMU extrinsics only. The final mean reprojection error is less than 0.5pixels.

3) Camera-LiDAR calibration: For the extrinsics of camera and LiDAR, we have developed a concise calibration toolbox based on 3D checker boards. We define the left RGB camera as center, then by sub-pixel extractions and extrinsics calculation, we can fully reconstruct the known-sized checkerboards to an accurate 3D model. At LiDAR side, we choose AVIA as reference because it works in non-repetitive scan mechanism which can integrate a dense point cloud in 1-2s. Then the two models are registered by point-to-plane ICP, and the camera-LiDAR extrinsics are thus solved, as illustrated in Fig. 4. The registration has achieved a precision of 9.1mm std.

4) Other calibrations: Based on the aforementioned process, an arterial camera-LiDAR-IMU calibration chain has already been established. The other sensors can either be calculated from the CADs, or be concatenated from the calibration chain. For example, AVIA manufacturer has provided explicit coordinates relation between LiDAR and its built-in IMU; Xsens and VLP16 also have explicit coordinates provided. To refine a better extrinsic for VLP16, we have performed a scan registration with AVIA, and the related params were updated in the chain. For the robot base, we have observed enough data from both the CADs and external measurements, achieving sub-cm calibration and have integrated it in the main chain, also.

## *E. Data Collection*

Our datasets are collected at 5th, 6th, 8th, and 18th of October, 2022 in a luxuriant botanic garden of our university. Various unstructured natural features are covered inside, such as thick woods, narrow trails, riverside, bridges, grasslands, as shown in Fig. 5. A total of 33 sequences are traversed, yielding 17.1 km trajectories, including short and long travels, cloudy and sunny days, loop closures, sharp turns, and monotonous textures, ideal for field robotic navigation research. Table III and Fig. 8 show the specifications and trajectories of 7 sample sequences which we have thoroughly tested with SOTA algorithms. The full dataset specifications can be accessed on our website.

## *F. Ground Truth Map*

Ground truth could be the most important part of a dataset. As indicated by Table I, most datasets fail to provide an authentic GT-map, which is necessary for evaluating the mapping results and plays a key role in robot navigation. To ensure the global accuracy, we have not used any mobile-mapping based techniques (*e.g.*, SLAM), instead we employ a survey-grade stationary 3D laser scanner and conduct a qualified surveying and mapping job with professional colleagues. The scanner is the RTC360 from Leica, which can output very dense and colored point cloud with a 130m scan radius and mm-level ranging accuracy, as shown the specifications in Table II. For possible future benefits, we have arranged two independent jobs both in early summer and middle autumn, which takes around 20 workdays in total, and respectively with 515 and 400 individual scans (each scan requires at least 3mins overall, Fig. 6 shows a work photo during the autumn survey). The scans are pre-registered by VI-SLAM and post-registered by Leica Cyclone Register360 software based on ICP and graph optimization (illustrated in Fig. 6). The final registered maps are generated in E57 format, and the coverage area is 48000m$^2$ from our calculation. According to the Leica Cyclone report, we have obtained an overall accuracy of 11mm across all possible links and loops within the map. This workflow is mature and trustable, as we have previously validated it using a total station, identifying an around 1cm global consistency for target points.

[3] https://linuxptp.sourceforge.net/
[4] https://www.mathworks.com/help/vision/camera-calibration.html
[5] https://github.com/ethz-asl/kalibr

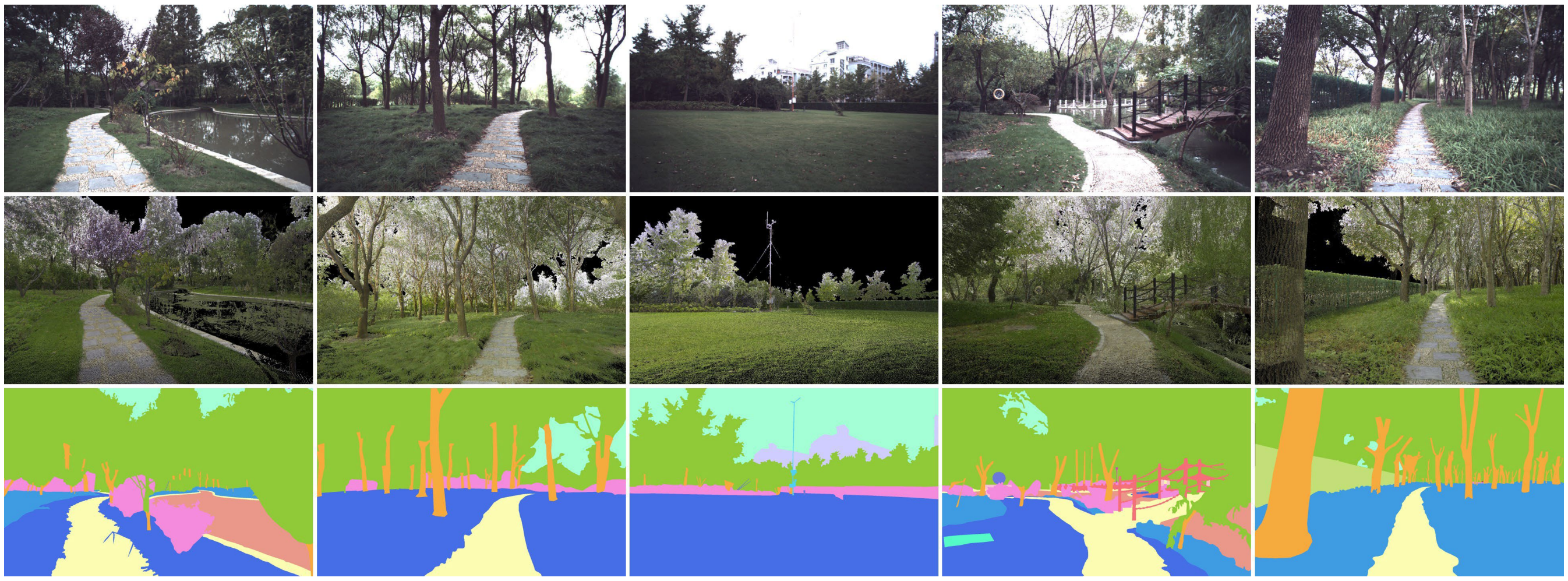

Fig.5. **Top**: Sample frames of typical scene features (riversides, thick woods, grasslands, bridges, narrow paths, *etc.*); **Middle**: The corresponding 3D map venues; **Bottom**: Dense semantic annotations of the corresponding frames.

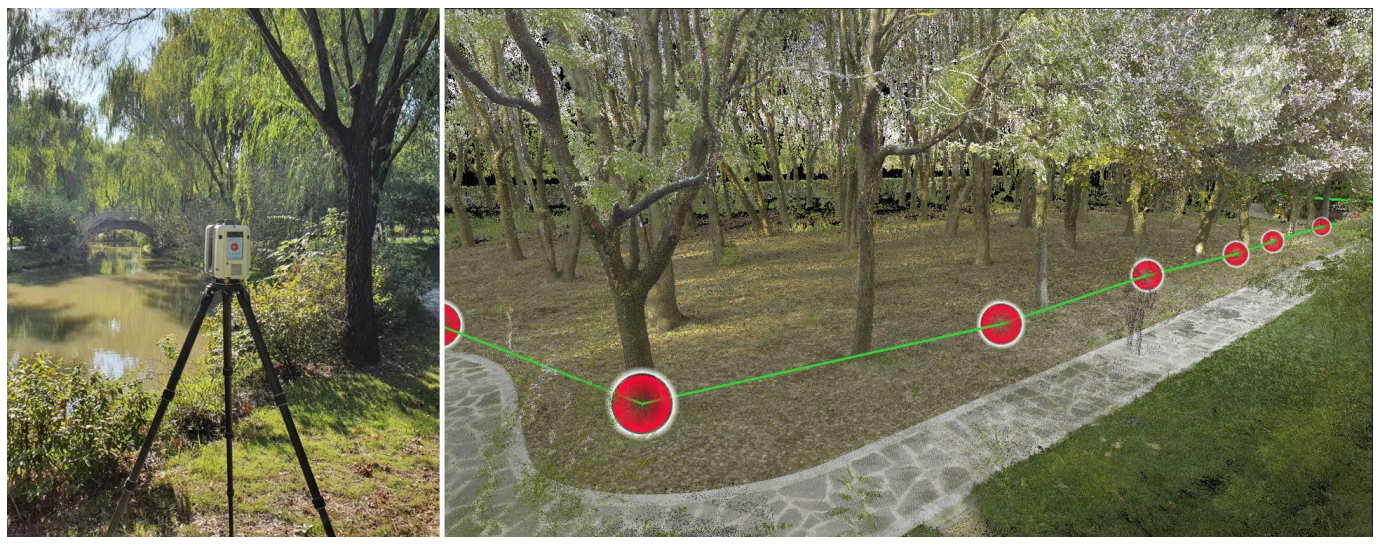

Fig.6. **Left**: The surveying process. **Right**: The point cloud registration process.

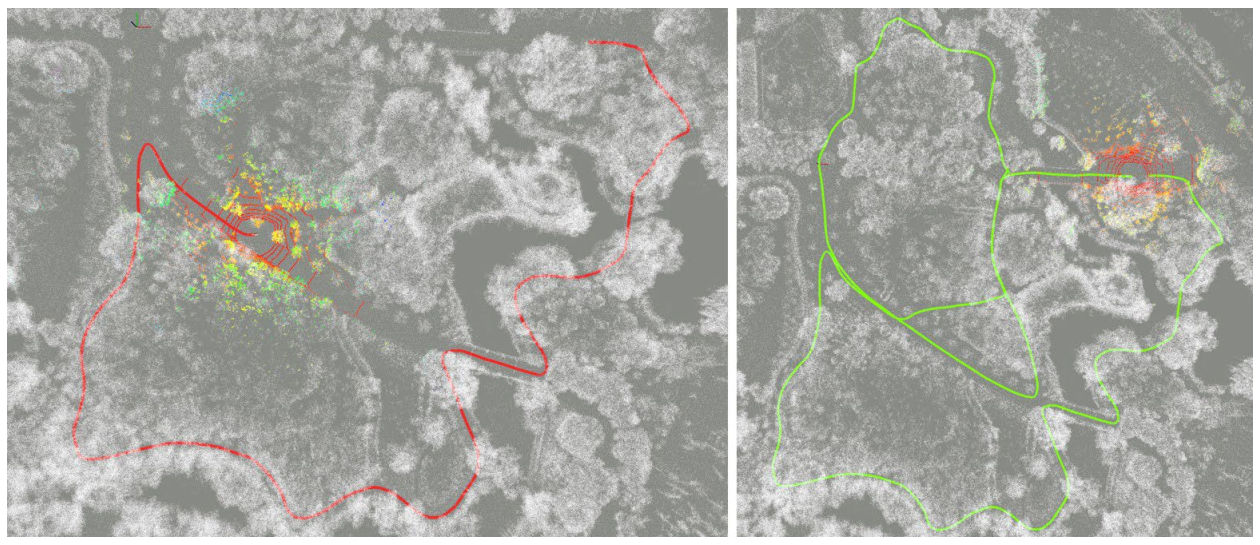

Fig.7. GT-pose generation based on our map-localization algorithm.

### G. *Ground Truth Pose*

Serving as the reference of navigation algorithms, ground truth poses are supposed to be complete and globally accurate. This is why GNSS is widely used for ground truth generation, while incremental techniques such as SLAM are not authentic due to the cumulative drift. However, in sky-blocked and complex environments, conventional means such as D-GNSS, Laser Tracker, and MoCap can hardly work consistently: our garden scenario exactly belongs to this scope. To bridge the gap, we take advantage of the authentic GT-map, and develop a map-based localization algorithm to calculate GT-pose using the on-robot VLP16 LiDAR. As the map is quite unstructured, and VLP16 is sparse, naive registration methods such as ICP cannot correctly converge on its own. This requires an accurate local tracking thread to provide a good initial pose for registration. To this end, we build a full-stack GT-pose system by fusing global initializer, VIO local tracker, and fine-registration modules. Firstly, the initializer searches the beginning frame in a scan-referenced image database for possible candidates, and subsequently an accumulated LiDAR segment can be registered to the GT-map for final initialization; Then, the VIO local tracker keeps estimating the robot motions to pre-register and undistort the LiDAR data; Finally, the fine-registration module employs point-to-plane ICP for final localization, as illustrated in Fig. 7. As the scene is really complex, we have slowed down the data playback rate and human monitored the visualization panel to make sure the GT-poses have converged correctly. To assess the accuracy of this method, we use a Leica MS60 laser tracker to crosscheck 32 stationary trajectory points in both normal- and dense-vegetated areas within the garden, resulting in 0.6cm and 2.3cm accuracy respectively. Even considering the LiDAR motion distortion that cannot be fully rectified (the up limit can be set by the 2-5% distance drift of LiDAR-inertial odometry [16], [41]), under an up to 15cm per frame motion speed, our GT-pose can still be defined with cm-level accuracy.

### H. *Semantic Annotation*

Semantic segmentation is the highest perception level of a robot. As a comprehensive and high-quality dataset, we emphasize the role that semantic information plays in navigation. Since our LiDARs are relatively sparse, we have arranged the annotation at 2D-image level. Our semantic segmentation database consists of 1181 images in total, including 27 classes such as various types of vegetations (bush, grass, tree, tree trunks, water plants), fixed facilities, drivable regions (trails, roads, grassland), rivers, bridges, sky, and more. The segmentation masks are meticulously generated with dense pixel-level human annotations, as shown in Fig. 5. All data are provided in LabelMe [42] format and support future reproductions. It is expected that these data can well facilitate robust motion estimation and semantic 3D mapping research. Additional detailed information can be accessed on our website.

TABLE III. SEQUENCES SPECIFICATIONS AND SOTA ALGORITHMS ASSESSMENT (VISUAL, LIDAR, AND SENSOR FUSION METHODS, SINGLE-RUN RESULTS)

| **Stat/Sequence** | **1005-00** | | **1005-01** | | **1005-07** | | **1006-01** | | **1008-03** | | **1018-00** | | **1018-13** | |
|---|---|---|---|---|---|---|---|---|---|---|---|---|---|---|
| Duration/s | 583.78 | | 458.91 | | 541.52 | | 738.70 | | 620.29 | | 131.12 | | 195.37 | |
| Distance/m | 601.60 | | 479.73 | | 591.04 | | 765.80 | | 750.09 | | 115.24 | | 201.23 | |
| Size/GB | 66.8 | | 49.0 | | 59.8 | | 83.1 | | 71.0 | | 13.0 | | 20.9 | |
| **Method/Metric [5]** | RPE/% | ATE/m | RPE/% | ATE/m | RPE/% | ATE/m | RPE/% | ATE/m | RPE/% | ATE/m | RPE/% | ATE/m | RPE/% | ATE/m |
| ORB3(stereo) [15] | X[1] | X | 5.586 N[2] | 5.933 N | X | X | 4.143 LC[3] | 3.453 LC | 4.148 LC | 5.005 LC | 5.220 N | 1.466 N | 5.303 N | 2.818 N |
| ORB3(stereo-imu) [15] | 4.386 N | 5.511 N | 4.808 N | 5.376 N | 4.771 N | 5.283 N | 3.733 LC | 3.150 LC | 3.853 LC | 4.311 LC | 4.118 LC | 1.116 LC | 4.238 N | 1.257 N |
| VINS-Mono [13] | 3.403 N | 8.617 N | 2.383 N | 4.029 N | 3.694 N | 7.819 N | 3.101 LC | 2.318 LC | 3.475 LC | 3.620 LC | 3.859 N | 1.767 N | 5.588 N | 2.967 N |
| LOAM [16] | 1.993 | 3.744 | 2.589 | 5.624 | 2.293 | 3.253 | 2.188 | 2.553 | 2.046 | 2.994 | 2.530 | 0.523 | 2.441 | 1.330 |
| Fast-LIO2 [41] | 1.827 | 2.305 | 1.870 | 2.470 | 2.349 | 4.438 | 6.573 | 39.733 | 2.404 | 4.019 | 2.770 | 2.154 | 2.562 | 2.390 |
| LVI-SAM [43] | 1.899 | 2.774 | 2.033 | 2.640 | 2.295 | 3.232 | 2.004 | 1.700 | 1.799 | 1.798 | 2.595 | 0.700 | 2.565 | 1.061 |
| R3LIVE [44] | 1.924 | 3.300 | 1.907 | 2.259 | 2.197 | 3.799 | 2.192 | 7.051 | 2.077 | 2.776 | 2.462 | 0.875 | 2.779 | 1.318 |

[1]X denotes the algorithm has tracking lost ≥20% of the sequence duration; [2]N denotes no loop closure detected; [3]LC denotes loop closure detected and corrected. Note that, LOAM and LVI-SAM are tested with VLP16 LiDAR, while Fast-LIO2 and R3LIVE are with AVIA LiDAR, which has a less horizontal field-of-view thus may perform weaker in degenerated venues.

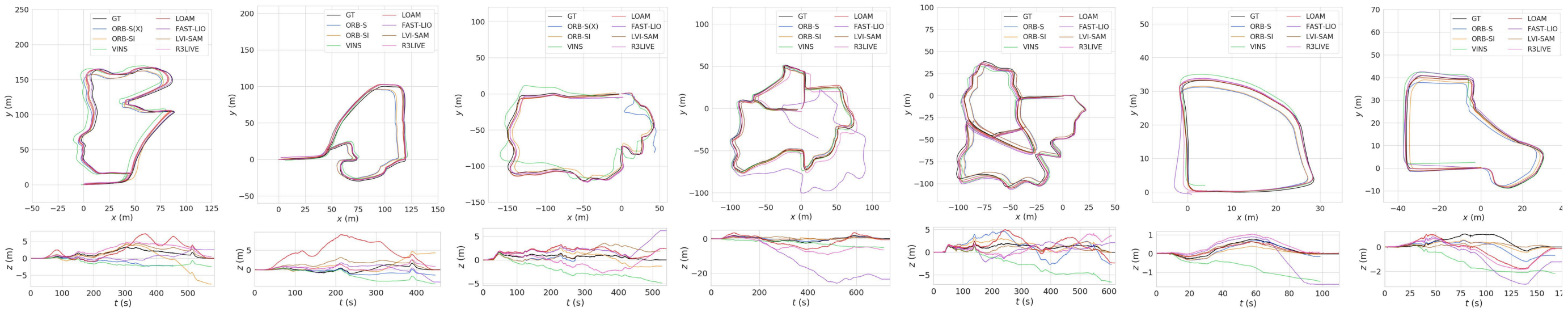

Fig.8. Visualization of the SOTA-estimated trajectories against the ground truth. From left to right: sequence 1005-00, 1005-01, 1005-07, 1006-01, 1008-03, 1018-00, and 1018-13; From top to bottom: top view of the trajectories, and the Z-axis errors.

## IV. EXAMPLE DATASET USAGE

### A. *Vision/LiDAR/Multi-sensor-fusion Navigation*

To verify the versatility of our dataset in navigation research, we select 7 sample sequences and conduct a thorough assessment on state-of-the-art algorithms (visual, LiDAR, and multi-sensor fusion methods) against the ground truth, regarding the metrics of relative pose error (RPE) and absolute trajectory error (ATE) [5]. The evaluation results are listed in Table III, and the trajectories comparisons are visualized in Fig. 8.

From the evaluation results we get mainly three conclusions:
1. Our dataset can support a wide range of navigation frameworks, including but not limited to stereo vision, visual-inertial, LiDAR-only, LiDAR-inertial, and visual-LiDAR-inertial based methods. This also demonstrates the good spatial calibration and time synchronization quality of our dataset.
2. Our dataset is a challenging benchmark for ground robots. As shown by the results, the RPE errors are around 5-10 times larger than KITTI leaderboard (ORB-stereo even failed 2/7 of the tests due to the indistinct textures and large view change at sharp corners); and it can be clearly identified that, most algorithms have met significant Z-axis error in the traverses, which should be paid more attention in future research. Besides, a noteworthy finding is that, although designed loop closures in all sequences, only 8/21 tests (visual methods) have succeeded in detection, indicating a high textural monotonicity of our data.
3. Multi-sensor fusion is an inevitable trend of future navigation research. It can be clearly seen that, compared with vision- and LiDAR-centric methods, multi-sensor fusion frameworks have earned very obvious elevation on both accuracy and robustness performance: we thus expect that our dataset can serve as a research incubator for novel sensor fusion mechanisms.

### B. *Other Possible Usage*

While our dataset is primarily designed for navigation research, its comprehensive data and ground truths enable its usefulness in various robotic tasks, including 3D mapping, semantic segmentation, image localization, depth estimation, *etc*. New chances and data will be continuously released on our website.

## V. CONCLUSION AND FUTURE WORK

This paper proposed *BotanicGarden*, a novel robot navigation dataset in problematic and unstructured natural environment involving GNSS denial, monotonous texture, and dense vegetations. In comparison to existing works, we have paid a lot of attention to dataset quality, incorporating comprehensive sensors, precise time synchronization, rigorous data loggings, and high-quality ground truth, all of which are at the top-level of this field. We firmly believe that our dataset can ease the research and inspire advancements for robot navigation.

In the future, we will continue to update and extend this dataset to enhance its complexity and comprehensiveness. Specifically, we will significantly increase the spatial coverage and trajectory length by traversing in and out of the garden. We will also arrange collections in various time periods and different weathers to append more challenges into the dataset. More importantly, we will enhance the platform by equipping it with independent GNSS receivers that are capable of providing raw satellite measurements. This will enable us to collect valuable source data throughout all four seasons under different vegetation states (from leaf-on to leaf-off) and satellite signal qualities, thereby supporting in-depth research and development towards novel and robust GNSS-integrated navigation systems.

## ACKNOWLEDGMENT

The authors would like to thank the colleagues from Tong Ji University and Sun Yat-sen University for their assistances in the rigorous survey works and post-processings, especially Xiaohang Shao, Chen Chen, and Kunhua Liu. We also thank A/Prof. Hangbin Wu for his guidance in data collection. Besides, we acknowledge Grace Xu from Livox for the support on Avia LiDAR, we acknowledge Claude Wu from Leica for the support on high-definition surveying, and we appreciate the colleagues of Appen for their professional works in visual semantic annotations. Yuanzhi Liu would like to thank Chenbo Gong for scene preparation work, and thank Jingxin Dong for her job-loggings and photographs during our data collection.